# Automated construction of sparse Bayesian networks from unstructured probabilistic models and domain information

Sampath Srinivas
Rockwell Science Center, Palo Alto Laboratory
444 High Street
Palo Alto, CA 94301

Stuart Russell
Computer Science Division
University of California
Berkeley, CA 94720

Alice Agogino
Department of Mechanical Engineering
University of California
Berkeley, CA 94720

## Abstract

An algorithm for automated construction of a sparse Bayesian network given an unstructured probabilistic model and causal domain information from an expert has been developed and implemented. The goal is to obtain a network that explicitly reveals as much information regarding conditional independence as possible. The network is built incrementally adding one node at a time. The expert's information and a greedy heuristic that tries to keep the number of arcs added at each step to a minimum are used to guide the search for the next node to add. The probabilistic model is a predicate that can answer queries about independencies in the domain. In practice the model can be implemented in various ways. For example, the model could be a statistical independence test operating on empirical data or a deductive prover operating on a set of independence statements about the domain.

## 1 Introduction

Bayes' belief networks (influence diagrams with only chance nodes) are usually constructed by knowledge engineers working with experts in the domain of interest. There are several problems with this approach. The expert may have only partial knowledge about the domain. In addition, there is a "knowledge acquisition bottleneck" problem when trying to build a knowledge base in this manner. It would be desirable to automate this modeling process such that belief networks could be constructed from partial domain information that may be volunteered by the expert and empirical data from the domain. Readers are referred to [3], [2], [5], [1] and [4] for details on Bayes' networks and influence diagrams.

### 1.1 A view of the general induction problem

The problem of inducing a Bayesian network from empirical data and domain information can be viewed as consisting of two subproblems:

1. How does one construct a dependency model for the variables in the domain? A dependency model is a set of statements of the form "$X$ is independent of $Y$ given $Z$" written as $I(X, Z, Y)$ where $X$, $Y$ and $Z$ are disjoint sets of variables in the model [3]. Thus, a predicate that can assign a truth value to statements of the form $I(X, Z, Y)$ is a dependency model.

2. Given a predicate of the form described above, how does one structure a sparse Bayesian network to represent the dependency model? There are various possible Bayesian network representations for the same dependency model. The problem is to construct a comprehensible and computationally efficient one.

The empirical data, for example, may be a collection of observations, each observation being a list of attribute value pairs (variable-value pairs) that represents a "snapshot" of the joint state of the domain variables. The domain information may consist of statements that can be used to infer facts about the dependence or independence relations among the variables in the domain (See Sec 1.2).

The solution to the first subproblem calls for a tractable statistical test for testing independence. The solution to the second subproblem requires building a structured model from an unstructured one. The work described in this paper concentrates on this structuring problem.

## 1.2 Problem statement

In the context of the previous section, a more precise statement of the problem we are solving here is as follows. We are given:

- A black box that can answer questions of the type "Is $X$ independent of $Y$ given $Z$?" where $X$, $Y$ and $Z$ are sets of variables in the domain. This could, for example, be a statistical test operating on empirical data or a deductive prover that knows the basic probability model axioms and operates on a declarative set of independence statements.

- Some partial expert information about the domain. The expert may make the following kinds of statements:
  - Declare that a variable is a hypothesis variable. Operationally, declaring that a variable $A$ is a hypothesis means that in the expert's view, $A$ is a root node of a belief network representation of the domain.
  - Declare that a variable is an evidence variable. Declaring a variable $A$ to be an evidence node means that the expert views $A$ as a leaf node in the belief network representation.
  - Declare that a variable $A$ is a cause of a variable $B$, or equivalently, that a variable $B$ is caused by variable $A$. Causality statements are interpreted as follows — Saying $A$ is a cause of $B$ declares that the expert views $A$ as a direct predecessor of $B$ in the belief network representation (see [3]).
  - Make explicit independence declarations of the form $I(X, Z, Y)$ where $X$, $Y$ and $Z$ are sets of domain variables.

Our goal is to build a sparse Bayesian network for the domain given the information above.

In a model it is usually easy for an expert to identify some 'primary' causes and some observables. The flow of causality in a causal model is from these primary causes to the observables. For example, in the medical domain, these primary causes are diseases and the observables are symptoms. In a model of a machine the primary causes would be possible faults and the observables would be sensors. Hypothesis variables correspond to primary causes and evidence variables to observables.

## 2 Bayesian networks

The work described in this paper makes use of the terminology and results found in Pearl [3]. A brief summary of the relevant material follows.

*Probabilistic models* comprise a class of dependency models. Every independence statement in a probabilistic model satisfies certain independent axioms (See [3] for details).

A belief network is a representation of a dependency model in the form of a directed acyclic graph (DAG). Given three disjoint node sets $X$, $Y$ and $Z$ in a directed acyclic graph, $X$ is said to be *d-separated* from $Y$ by $Z$ if there is no *adjacency* path from $X$ to $Y$ that is *active*. An adjacency path follows arcs from a node in $X$ to a node in $Y$ without regard to the directionality of the arcs. An adjacency path from $X$ to $Y$ is active if (1) if every node in the path that has converging arrows [1] is in $Z$ or has a descendant in $Z$. (2) Every other node is outside $Z$. We represent the statement "$X$ is d-separated from $Y$ by $Z$" as $D(X, Z, Y)$.

A DAG is called an Independency map (*I-map*) if every d-separation in the graph implies the corresponding independence in the underlying dependency model that the DAG is attempting to represent, i.e:

$$D(X, Z, Y) \Longrightarrow I(X, Z, Y) \qquad (1)$$

A belief network is called a dependency map (*D-map*) if every non-d-separation in the graph implies a corresponding non-independence in the underlying dependency model, i.e:

$$\neg D(X, Z, Y) \Longrightarrow \neg I(X, Z, Y) \qquad (2)$$

---

[1] A converging arrows node in an adjacency path is a direct successor (in the DAG) to its neighbours in the path.

Or, equivalently:

$$I(X, Z, Y) \Longrightarrow D(X, Z, Y) \quad (3)$$

If a DAG is both an I-map and a D-map it is called a *perfect map* of the dependency model. DAGs cannot represent all the possible kinds of independencies in dependency models. In other words, for many dependency models, there is no DAG that is a perfect map. Therefore, if a DAG representation is used for such a model, one that shows a maximum amount of useful information about the model should be chosen. A DAG that shows no spurious independencies while explicitly displaying as many of the model's independencies as possible is a reasonable compromise. Such a DAG is called a *minimal I-map*. Deletion of any edge of a minimal I-map makes it cease to be a I-map. It is to be noted that a dependency model may have many different minimal I-maps.

Let $M$ be a dependency model and $d = X_1, X_2, \ldots X_N$ be an ordering defined on the variables of the model. The boundary stratum $B_i$ of the node $X_i$ is defined to be a minimal subset of $\{X_1, X_2 \ldots X_{i-1}\}$ such that $I(\{X_i\}, B_i, \{X_1, X_2 \ldots X_{i-1}\} - B_i)$. The DAG created by designating the nodes corresponding to the variables in $B_i$ as the parents of the node corresponding to $X_i$, for all $i$, is called a boundary DAG of $M$ relative to the node ordering $d$. If $M$ is a probabilistic model then a boundary DAG of $M$ is a minimal I-map. A *Bayesian network* is a minimal I-map of a probabilistic model.

The result above is a solution to the problem of building a *Bayesian network* for a given probabilistic model and node ordering $d$. The form of the Bayesian network depends strongly on the order of introduction $d$ of the variables.

In the Boundary DAG algorithm the particular ordering of nodes chosen can make a large difference in the number of arcs in the resulting Bayesian network. Though the resulting belief network is guaranteed to be a minimal I-map, this does not imply that it is sparse. For example, take the networks in Figure 1[2].

Fig 1a is a perfect map of a probability model that describes the dependence of two temperature sensors $T1$ and $T2$ on an (unobservable) process temperature $T$. If the boundary DAG algorithm is used to build a Bayesian network from the underlying probability model with node ordering $d =$ $\{T, T1, T2\}$ the resulting Bayesian network is the same as the perfect map Fig 1a. If the node ordering $d = \{T2, T1, T\}$ is used instead, we get Fig 1b. Though this network is a minimal I-map, it is fully connected and carries no information on conditional independence. Fig 1a can be viewed as a causal network in which the fact that the hypothesis node $T$ makes the evidence nodes $T1$ and $T2$ conditionally independent is explicitly recorded.

---

[2] This example is adapted from [4].

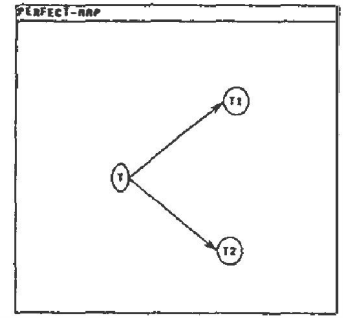

(a)

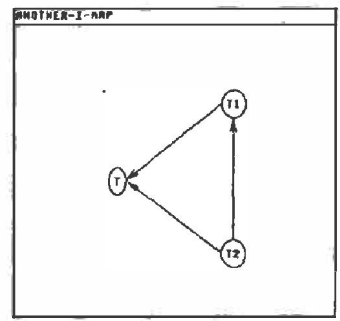

(b)

Figure 1: Different minimal I-maps for the same probability model

A belief network has to be sparse if it is to be comprehensible to the user and inference using the network is to be computationally tractable. Using the boundary DAG algorithm as a point of departure, our solution to the problem of building a belief network from a probabilistic model and expert information attempts to build a sparse Bayesian network by choosing an appropriate ordering of the nodes.

# 3 The construction algorithm

The boundary DAG method [3] is a simple way of building a Bayesian network for a set of variables. The recursive statement of the algorithm for building a Bayesian network of $k + 1$ variables is as follows:

> Given: A Bayesian network $K$ consisting of $k$ variables and a variable $X_{k+1}$ to be added to the Bayesian network.
>
> Algorithm: Using the independence predicate find the smallest subset $P$ of the variables in $K$ such that $I(\{X_{k+1}\}, P, K - P)$. Designate the variables in $P$ as the predecessors of $X_{k+1}$.

We could adapt this algorithm to build a *sparse* Bayesian network if we could choose $X_{k+1}$ in a coherent way from all the variables which have not yet been added to the network. This is like a search problem. If there are $n$ nodes in all, there are $n - k$ nodes left to add at each recursive step of the algorithm. The problem is to find the best one to add. Ideally what we would like to do is to find the most sparse minimal I-map, i. e., among the $n!$ possible minimal I-maps (Bayesian networks) possible for $n$ nodes using the Boundary DAG algorithm (one for each ordering of nodes), find the one that has the least number of arcs. This is possible, in principle, with a complete search. The complexity of such a procedure is prohibitive.

The algorithm we have implemented for choosing node $X_{k+1}$ uses a *priority* heuristic based on the expert's information and a greedy sub-heuristic to make the choice. The priority heuristic ensures that hypothesis nodes are added to the network before evidence nodes and that cause nodes are added before effect nodes, thus guiding the algorithm towards a sparse causal network. If there is not enough information to make a choice based on priority the node which adds the least number of arcs to the existing network is chosen as node $X_{k+1}$.

The actual implementation of the priority heuristic is as follows. The expert's information is first 'compiled' into a DAG. *Cause-of* and *caused-by* relations are translated into appropriate directed links between nodes in the DAG. Nodes which are declared to be *hypothesis* or *evidence* nodes are annotated as such. This DAG is distinct from the belief network being constructed by the algorithm and is used solely for the purpose of making priority decisions.

Priority is a function that defines a partial ordering among the nodes in the expert information DAG. The relative priority between two nodes $A$ and $B$ is decided as follows: If $A$ is a hypothesis node and $B$ is not a hypothesis node $A$ has higher priority. If $A$ is an evidence node and $B$ is not an evidence node $A$ has lower priority. If $A$ is an ancestor of $B$ in the expert information DAG it has higher priority. If $A$ is a descendant of $B$ then $A$ has lower priority. If none of the above cases apply the priority ranking of $A$ and $B$ is same.

The recursive algorithm used for building a sparse Bayesian network of $k + 1$ variables is now stated as follows:

> Given: A Bayesian network $K$ with $k$ nodes and $n - k$ candidate nodes which are yet to be added to the network.
>
> Algorithm:
>
> 1. Order the candidates using the *priority* ordering. If there is a unique candidate with highest priority choose it as the winner, i.e., the next node that will be added to the network.
>
> 2. If there is a set of candidates with the (same) highest priority, find the boundary stratum of each of the candidates and choose the candidate with the smallest boundary stratum as the winner. This is a greedy heuristic that tries to minimize the number of arcs added.
>
> 3. If there is still a tie choose any candidate as the winner from the remaining candidates.
>
> 4. Make the winner $X_{k+1}$, the $k + 1$th node in the network. Find the winner's boundary stratum if it has not been found already. Assign the boundary stratum of the winner as predecessors of $X_{k+1}$ in the belief network being constructed.

The boundary stratum of a candidate $C$ is found by generating all possible subsets of nodes $S$ of the nodes in the existing diagram $K$ in increasing order of size until a subset $S_c$ is found which satisfies $I(C, S_c, K - S_c)$. The order of generation of the subsets guarantees that $S_c$ is the smallest subset of $K$ that satisfies the above independence condition. In other words $S_c$ is the boundary stratum.

The algorithm recurses until all the nodes have been added to the network.

### 3.1 Complexity

The algorithm outlined above requires $(n-k)2^k$ independence checks when adding the $k+1$th node. The total number of independence checks required is $O(2^{n+1})$. Using the contrapositive form of the decomposition axiom for probabilistic models [3] it can be shown that once a particular subset $S$ of a partial belief network $K$ has been found *not* to be a boundary stratum for a candidate node $C$ it will not be found to be a boundary stratum for the candidate $C$ even if the belief network $K$ is augmented with some new nodes. This allows us reduce the total number of independence checks to $O(2^n)$. Nevertheless, the algorithm is still exponential.

Despite this, it should be noted that once a boundary stratum $S_c$ for a candidate node $C$ has been found, there is no need to check all the subsets of $K$ which are larger than $S_c$. The hypothesis in having the expert information available is to guide the algorithm along towards a sparse belief network. If this is indeed the case, the $S_c$'s are small and the algorithm runs in far smaller than exponential time. For example, if we operationalize our sparseness hypothesis as the assumption that the maximum number of predecessors that a node can have is $p$, then at each step of the construction algorithm we need to check only those subsets of nodes of the existing network which are of size less than or equal to $p$ to find the boundary stratum of a candidate node. The overall complexity in such case is polynomial ($O(n^{p+2})$).

Indeed, tests with the implemented system show that the algorithm takes far less time and generates results that are more desirable (belief nets with smaller number of arcs) as the amount of expert information available increases. In the trivial extreme, it is possible for the expert to basically "give the system the answer" by giving enough information to allow the total ordering of the nodes. In such a case the system serves to verify the expert's intuitions rather than to fill in the gaps in the expert's model.

## 4 Results

The belief network construction algorithm described above has been implemented in Common Lisp on a Symbolics workstation. The system is an experimental module in a comprehensive belief network and influence diagram evaluation and analysis package called IDEAL [6].

During testing of the system the underlying probability model has been implemented as a belief network and the independence test has been implemented as d-separation in this belief network. Though this may seem a little strange at first, it should be borne in mind that we are concentrating on the *construction* of a sparse belief network given a probability model and an independence predicate that operates on the model (Sec 1.1). The exact implementation of the model and the test do not have any effect on the results of the algorithm. In addition, there is a benefit to testing the system this way. The topology of the underlying belief network (i.e the belief network that represents the underlying probability model) gives us a standard against which our rebuilt network can be compared. The best possible rebuilt network will be identical to the underlying network since, in that case, it is a perfect map of the underlying model. All other possible rebuilt networks will be minimal I-maps of the underlying belief network but will not necessarily show all the independencies embodied in the underlying network.

During the construction of the expert information DAG obvious contradictions in the expert's information are detected by the system and have to be corrected by the user. For example, the expert may specify a set of *cause-of* relations that lead to a cycle in the DAG. Another example of an error is specifying that a hypothesis node is caused by some other node. The system also verifies the expert's information as it builds the network and warns the user when it finds deviations between the model it is building and the expert's information. For example, the expert may have declared that variable $A$ is a cause of variable $B$ while the system may find that the boundary stratum of $B$ does not contain $A$.

Fig 2 is an example of an underlying network, the expert information (as a partially specified DAG) and the rebuilt network. The expert information consists of (1) the identities of all the evidence nodes ($Y3$, $Y2$ and $Y1$) and hypothesis nodes ($U2$ and $U1$). (2) knowledge of the existence of some arcs (see Fig 2b). The rebuilt network is similar to the underlying network except for the arcs among the subset of nodes $W1$, $W2$, $U1$ and $V$. It is interesting to note that the rebuilt network can be obtained from the original by two sequential arc reversals — reversal of arc $V \longrightarrow W2$ followed by reversal of arc $V \longrightarrow W1$ [3].

If the arc $V \longrightarrow W2$ is added to the expert information DAG then the rebuilt network is identical to the underlying network (Fig 3).

Fig 4 is a slightly larger example. Here we have attempted a crude calibration of the sensitivity of the system to the amount of causal expert information available. The underlying belief network has 26 nodes and 36 arcs. The expert information initially consists of (1) the identities of all the evidence and hypothesis nodes and (2) knowledge of the existence of all the arcs (i.e the expert knowledge consists of causal statements that describe all the arcs in the underlying diagram). The system builds the network with this information. The knowledge of the existence of some random arc in the underlying model is then deleted from the expert information DAG. The system builds the network again. This delete/build cycle is repeated many times. Figure 5a shows the variation in the number of arcs in the rebuilt network versus the number of arcs in the expert information DAG. Taking the number of arcs in the rebuilt network to be a rough indicator of the quality of the rebuilt network we see that the quality of the model improves as the amount of expert information available increases. Figure 5b shows the amount of time required to build the model against the number of arcs in the expert information DAG. The time required decreases as the expert information available increases.

## 5 Discussion and further work

As expected, the belief network constructed by the system depends strongly on the amount of expert information available. We are at present trying to characterize what types of information are critical to building good models. Our experience with the system shows that the identification of hypothesis and evidence nodes by the expert seems very important if a reasonable model is to be built.

If this system is to be applied to induce a belief network from empirical data it is imperative that an inexpensive and fairly accurate statistical independence test be used. Well characterized conditional independence tests involving larger numbers of variables may not be tractable. It may be necessary, therefore, to make use of appropriate approximation techniques or less formal tests that may be more

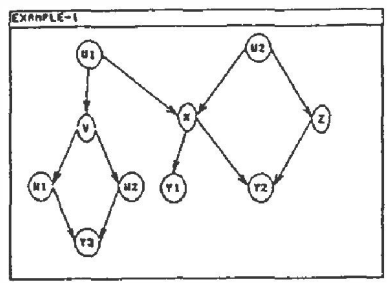

a) Underlying network

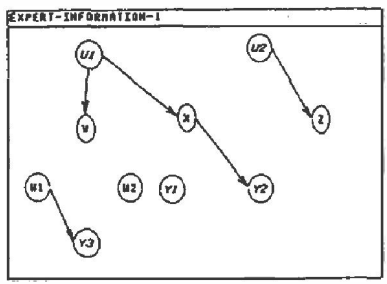

b) Expert information

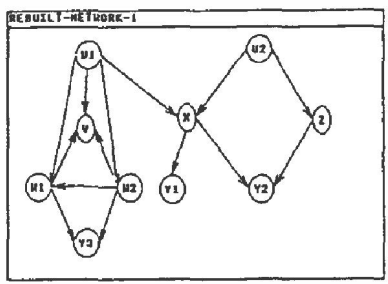

c) Rebuilt network

Figure 2: An example

---
[3] See [5] for details on arc reversals.

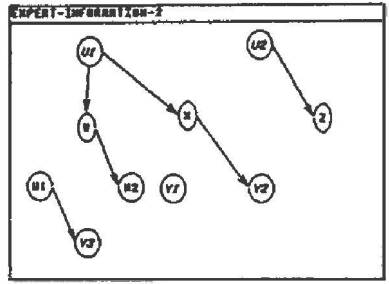

a) Expert information

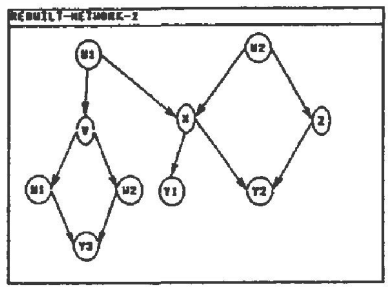

b) Rebuilt network

Figure 3: Rebuilt network with additional expert information

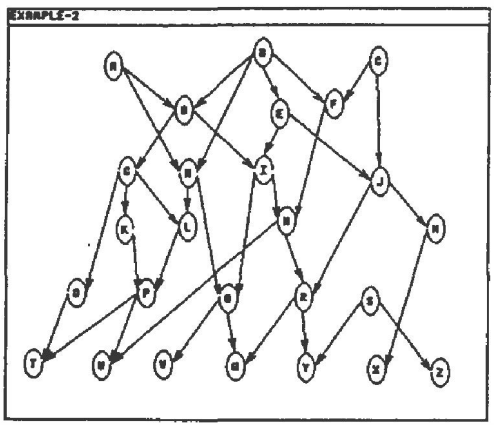

Figure 4: A larger example

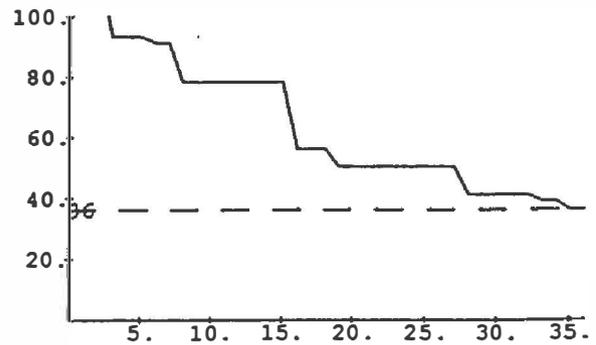

X: Number of arcs in expert information
Y: Number of arcs in rebuilt network
(a)

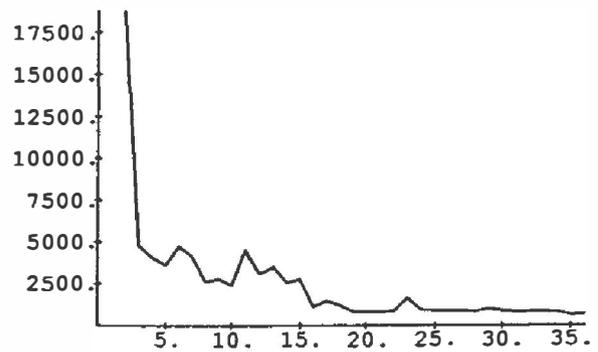

X: Number of arcs in expert information
Y: Time taken to build network (seconds)
(b)

Figure 5: Sensitivity to amount of expert information

tractable. An additional and more basic 'problem' with statistical independence tests on empirical data is that they can never be exact. This fact can be regarded as a characteristic of the induction problem. In this regard, it is interesting to note that a Bayesian network and d-separation provide a sound and complete scheme to deduce, in polynomial time, every independence statement that is implied by the Bayesian network [3]. This property made a Bayesian network and d-separation an attractive scheme to use to represent the underlying model and independence test during testing of the system.

The system, as it is implemented now, guarantees that the network constructed is a Bayesian network, i.e a minimal I-map. The expert information is used merely to guide the search for the next node. The actual boundary stratum of the next node is found from scratch using the independence test. Thus, in a sense, the system "trusts" the underlying model more than the expert. Substantial computational gains could be achieved if the system started out looking for the boundary stratum of a node under the assumption that the causes that the expert has declared for the node are necessarily members of the boundary stratum of the node. However, using this approach would no longer guarantee that the belief network constructed is a minimal I-map of the underlying model.

# 6 Acknowledgements

We would like to thank Jack Breese and Ken Fertig for their invaluable help and advice.